\documentclass{article}

\usepackage{authblk}
\usepackage{amsfonts}
\usepackage[version=3]{mhchem}
\usepackage{textgreek}
\usepackage{subcaption}

\usepackage{booktabs, caption, makecell}

\usepackage{threeparttable}
\usepackage{pdflscape} % landscape pages
\usepackage{lmodern} % Might prevent warnigns about unabailable fonts
\usepackage[margin=1in]{geometry} % bigger margins
\date{}
\usepackage{hyperref}
\hypersetup{
    colorlinks=true,
    linkcolor=blue,
    filecolor=magenta,      
    urlcolor=cyan,
}

\title{BanglaLekha-Isolated: A Comprehensive Bangla Handwritten Character Dataset}

\author[1]{Mithun Biswas}
\author[1]{Rafiqul Islam}
\author[1]{Gautam Kumar Shom}
\author[2]{Md. Shopon}
\author[1]{Nabeel Mohammed}
\author[1]{Sifat Momen}
\author[1]{Md Anowarul Abedin}
\affil[1]{Department of Computer Science and Engineering, University of Liberal Arts Bangladesh}
\affil[2]{Department of Computer Science and Engineering, University of Asia Pacific}

\usepackage{Sweave}
\begin{document}
\maketitle

\begin{abstract}
Bangla handwriting recognition is becoming a very important issue nowadays. It is potentially a very important task specially for Bangla speaking population of Bangladesh and West Bengal. By keeping that in our mind we are introducing a comprehensive Bangla handwritten character dataset named BanglaLekha-Isolated. This dataset contains Bangla handwritten numerals, basic characters and compound characters. This dataset was collected from multiple geographical location within Bangladesh and includes sample collected from a variety of aged groups. This dataset can also be used for other classification problems i.e: gender, age, district. This is the largest dataset on Bangla handwritten characters yet.  
\end{abstract}
%%%%%%%%%%%%
% Intro
%%%%%%%%%%%%
\section{Introduction}

Bangla is the mother language of
Bangladesh and the 7th most widely spoken language in the world. There are more than 200 million native Bangla speakers. It is the official language of Bangladesh and several Indian states including West Bengal, Tripura, Assam and Jharkhand. Bangla is also the official language of Sierra Leone a West African country. With the rapid adoption of technology in different sectors in these regions, recognizing handwritten Bangla characters is an important challenge to overcome. While there has been great successes in the application of machine learning tools for the English language, the same level of effectiveness is not observed for Bangla. One of the many reasons for this is the lack of a single comprehensive dataset which covers the frequently used Bangla characters. There are existing data sets which cover either just the Bangla numerals, or just the Bangla characters, or just the Bangla compound characters. While it is possible to combine them to form a unified data set, the incovenience faced by the researchers stem from the lack of consistency in the data presentation of the different data sets. 
BanglaLekha-Isolated is the first of a chain of datasets being released which aims to foster Bangla handwriting related research by:

\begin{itemize}
	\item Providing a large dataset suitable for machine learning applications which include the most frequently used Bangla characters covering Bangla numerals, basic characters and compound characters.
    \item Provide a suitably pre-processed version of the dataset to reduce the time between data set acquisition and reporting results.
    \item Provide multiple labels per character/character group to facilitate research in: 
    \begin{itemize}
    	\item Automatic recognition certain characteristics of the writer (Age, gender, location etc)
        \item Automatic assessment of handwriting quality and  methods of giving useful feedback.
	\end{itemize}
    	
\end{itemize}

The BanglaLekha-Isolated dataset contains smaples of 84 different Bangla handwritten numerals, basic characters and compound characters. A comparison with the two other popular sources of Bangla handwriting related datasets (CMATERDB\cite{CMATERDB} and the ISI Handwriting datasets\cite{ISI}) are given in Table \ref{TABLE}.

\begin{figure}
\centering
\begin{subfigure}{.2\textwidth}
  \centering
  \includegraphics[width=.9\linewidth]{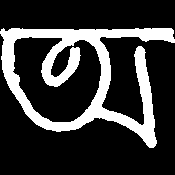}
  \caption{Basic}
  \label{fig:sub1}
\end{subfigure}%
\begin{subfigure}{.2\textwidth}
  \centering
  \includegraphics[width=.9\linewidth]{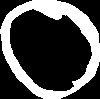}
  \caption{Numeral}
  \label{fig:sub2}
 \end{subfigure}
\begin{subfigure}{.2\textwidth}
  \centering
  \includegraphics[width=.9\linewidth]{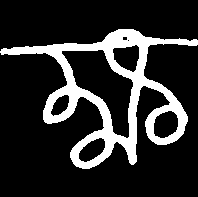}
  \caption{Compound}
  \label{compound}
 \end{subfigure}
\caption{Sample Images of BanglaLekha-Isolated}
\label{fig:test}
\end{figure}

The BanglaLekha-Isolated dataset consists total of 166,105 square images (while preserving the aspect ratio of the characters), each containing a sample of one of 84 different Bangla characters. The number of samples in each class are almost equal, which is not the case in some of the other datasets (e.g. CMATERDB Compound character set). The 84 characters classes contain 10 numerals, 50  basic characters and 24 frequently used compound characters. Some samples images of the dataset are shown in Figure \ref{fig:test}.

\begin{table}
\caption{Number of images in different datasets}
\label{TABLE}
\begin{center}
\begin{tabular}{ |p{3cm}||p{3cm}|p{3cm}|p{3cm}|  }
 \hline
Type & CMATERDB\textsuperscript{1}\cite{CMATERDB} &ISI Dataset\textsuperscript{2} \cite{ISI}&BanglaLekha-Isolated\\
 \hline
 Basic Character   & 15,103    &30,966&   98,950\\
 \hline
 Numerals &   6,000  & 23,299   &19,748\\
 \hline
 Compound Characters & 42,248 & None &  47,407\\
 \hline

\end{tabular}
\end{center}
\begin{tablenotes}
            \item[a] \textsuperscript{1} CMATERDB dataset has 3 different datasets for basic characters,numerals and compound characters.
            \item[b] \textsuperscript{2} ISI dataset has two different dataset for basic characters and numerals.
\end{tablenotes}
\end{table}

\section{Data Collection and Pre-processing}
\begin{figure}
\centering
\includegraphics{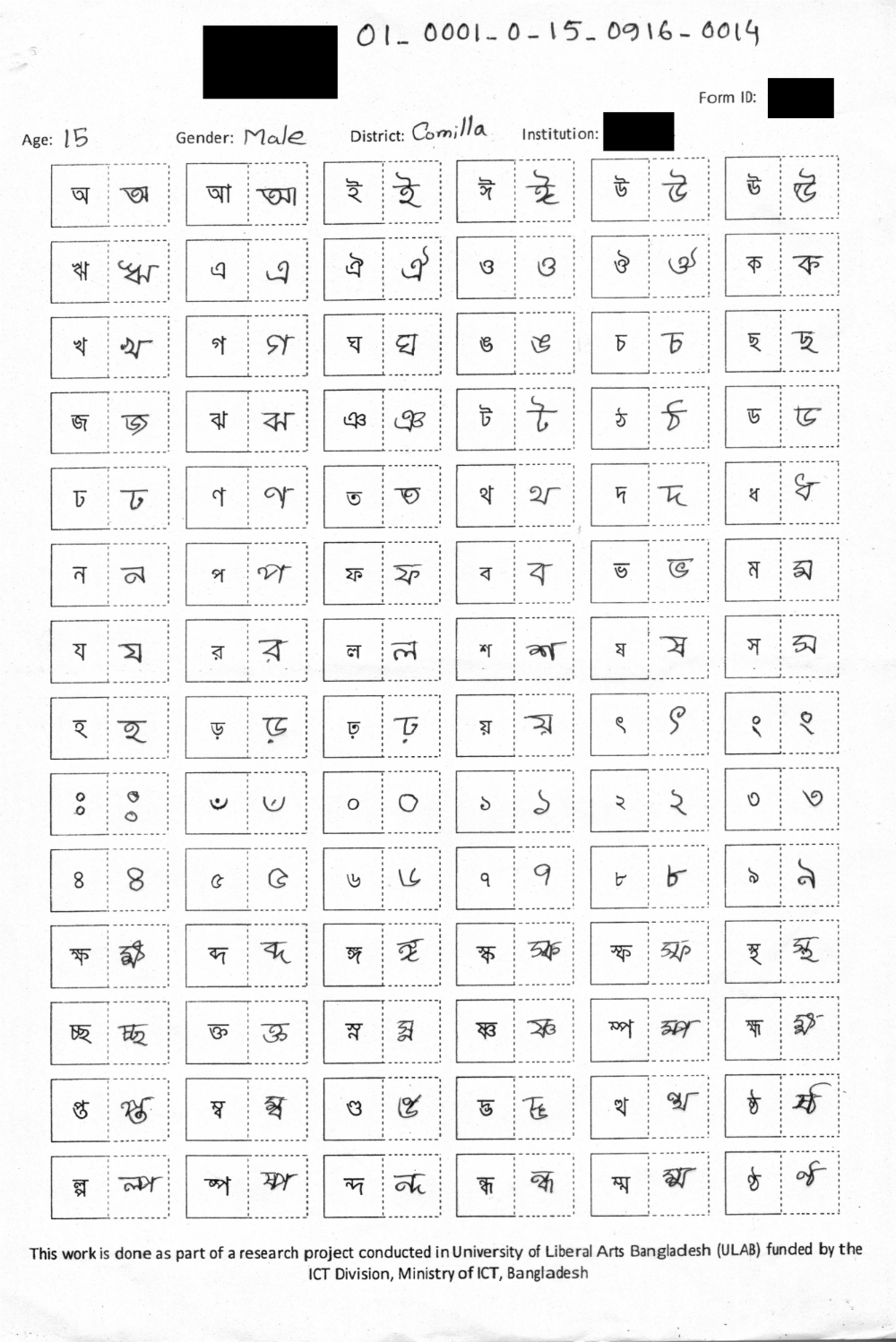}
\caption{Form that was used for collecting dataset. \label{p_map}}
\end{figure}

This dataset was collected from literate native Bangla speakers of different regions and with age range between 4 to 27. A small fraction of the samples were collected from individuals with physical disabilities. Each individual was supplied with a form similar to the one shown in Figure \ref{p_map}. For a wider distribution of handwriting quality, samples were collected  specific time constraints (10 Minutes, 5 Minutes, 2 Minutes). Each subject also gave information about her/his age, gender, and district he lives in.

The images that are present in the dataset were pre-processed in the following ways:
\begin{itemize}
  \item Foreground and background were inverted so that images have a black background with the letter drawn in white.
  \item Noise removal was attempted by using the median filter.
  \item An edge thickening filter was applied.
  \item Images were resized to be square in shape with appropriate padding applied to preserve the aspect ratio of the drawn character.
\end{itemize}

\section{Possible Uses of BanglaLekha-Isolated Dataset}
Our dataset can be used for handwritten character recognition, which is obvious, but there are some more features that can be used for research purpose using our dataset. As it is already mentioned in Section 1 that it is possible to work on automatic recognition on certain characteristics of the writer such as age,gender, location etc. These informations can be used even for forensic purposes.  
\section{Naming Convention}
Each and every sample of BangLekha-Isolated dataset has an unique form ID by which the age, gender, district, and Institute of the participants can be identified. So, a 22 characters long form Id was proposed, where first 2 digit is for the district, then the next four digits is for Institution, the next one digit is for gender and the next two is for age, again the next four is for date and the last four is used for form serial number and every information (digit part) is separated by an underscore. For example- $$ \textbf{01\_0001\_0\_09\_1016\_0001}$$ is a unique form id and here 01 means the participant is form Comilla, 0001 means participant is from Ispahani School and College, then 0 means the participant is a male and 1016 means the participant filled up the form in October of 2016 and 0001 is the form serial number. So whenever one used any character form this dataset (around 1,68,000 data), he/she can get the information (age, gender, district, etc.) of the participants. 

\section{Marking}
All the 2000 forms that were collected were marked by 3 native Bangla speakers using the criteria set by a handwriting specialist. The judgment on the mark is based on:
\begin{itemize}
\item Shape of the characters
\item Clarity of the image
\item Appropriate use of matra (A horizontal straight line put over the consonants and some vowels of the Bengali alphabet)
\item Subjective evaluation based on beauty of letters
\end{itemize}

The marks are also provided with the dataset in a separate spreadsheet.

\section{Conclusion}
BanglaLekha-Isolated dataset aims for creating new scopes for researchers who are interested in working on Bangla handwritten characters. The dataset is available in \cite{BANGLALEKHA} \href{http://www.banglalekha.org}. This report documents the initial release of the data set. As more refinements are done and/or new data sets are collected, this report will be updated as appropriate.

\addtolength{\textheight}{-2.725cm}


\begin{thebibliography}{6}
\bibitem{CMATERDB} "Cmaterdb - CMATERdb: The pattern recognition database repository," http://www.findbestopensource.com/product/cmaterdb, accessed: 2017-02-20.
\bibitem{ISI} U. Bhattacharya and B. B. Chaudhuri, "Handwritten numeral databases of indian scripts and multistage recognition of mixed numerals," IEEE transactions on pattern analysis and machine intelligence, vol. 31, no. 3, pp. 444-457, 2009.
\bibitem{BANGLALEKHA} M. Biswas, R. Islam, G. Kumar, M. Shopon, N. Mohammed, S. Momen, and A. Abedin, "Banglalekha-isolated: A comprehensive bangla handwritten dataset." [Online]. Available: http://www.banglalekha.org/dataset
\end{thebibliography}
\end{document}